\documentclass[10pt,twocolumn,letterpaper]{article}

\usepackage{iccv}
\usepackage{times}
\usepackage{epsfig}
\usepackage{graphicx}
\usepackage{amsmath}
\usepackage{amssymb}

\usepackage{epsfig}
\usepackage{graphicx}
\usepackage{amsmath}
\usepackage{amssymb}
\usepackage{multirow}
\usepackage{enumerate}
\usepackage{bm}
\usepackage{wrapfig}
\usepackage{epstopdf}

\usepackage[breaklinks=true,bookmarks=false]{hyperref}

\iccvfinalcopy 

\def\iccvPaperID{1173} 

\ificcvfinal\fi

\begin{document}

\title{Unsupervised Learning of Fine Structure Generation for 3D Point Clouds by 2D Projection Matching}

\author{Chao Chen$^\ast$\\
School of Software, BNRist, Tsinghua University\\
{\tt\small chenchao19@mails.tsinghua.edu.cn}
\and
Zhizhong Han\thanks{indicates the equal contribution. This work was supported by National Key R$\&$D Program of China (2020YFF0304100), the National Natural Science Foundation of China (62072268), and in part by Tsinghua-Kuaishou Institute of Future Media Data, and NSF (award 1813583). The corresponding author is Yu-Shen Liu.}\\
Wayne State University\\
{\tt\small h312h@wayne.edu}
\and
Yu-Shen Liu\\
School of Software, BNRist,
Tsinghua University\\
{\tt\small liuyushen@tsinghua.edu.cn}
\and
Matthias Zwicker\\
University of Maryland, College Park\\
{\tt\small zwicker@cs.umd.edu}
}

\maketitle
\ificcvfinal\thispagestyle{empty}\fi

\begin{abstract}
   Learning to generate 3D point clouds without 3D supervision is an important but challenging problem. Current solutions leverage various differentiable renderers to project the generated 3D point clouds onto a 2D image plane, and train deep neural networks using the per-pixel difference with 2D ground truth images. However, these solutions are still struggling to fully recover fine structures of 3D shapes, such as thin tubes or planes. To resolve this issue, we propose an unsupervised approach for 3D point cloud generation with fine structures. Specifically, we cast 3D point cloud learning as a 2D projection matching problem. Rather than using entire 2D silhouette images as a regular pixel supervision, we introduce structure adaptive sampling to randomly sample 2D points within the silhouettes as an irregular point supervision, which alleviates the consistency issue of sampling from different view angles. Our method pushes the neural network to generate a 3D point cloud whose 2D projections match the irregular point supervision from different view angles. Our 2D projection matching approach enables the neural network to learn more accurate structure information than using the per-pixel difference, especially for fine and thin 3D structures. Our method can recover fine 3D structures from 2D silhouette images at different resolutions, and is robust to different sampling methods and point number in irregular point supervision. Our method outperforms others under widely used benchmarks. Our code, data and models are available at \href{https://github.com/chenchao15/2D\_projection\_matching}{https://github.com/chenchao15/2D\_projection\_matching}.
\end{abstract}

\section{Introduction}
It is important to learn to generate 3D point clouds in different 3D computer vision applications, such as single image reconstruction~\cite{InsafutdinovD18,Navaneet2019,navaneet2019differ,handrwr2020,Yifan:DSS:2019a} and novel shape generation~\cite{MAPVAE19,ShapeGF,DBLP:journals/corr/abs-1906-12320,Pumarola_2020_CVPR}. The latest supervised methods~\cite{Fan_2017_CVPR,cvprpoint2017,nipspoint17,p2seq18,MAPVAE19} leverage deep neural networks to learn to generate 3D point clouds from latent codes using 3D ground truth. However, it is expensive and tedious to obtain large scale 3D ground truth data sets, which significantly affects the supervised learning performance.

Unsupervised methods~\cite{InsafutdinovD18,lin2018learning,Navaneet2019,navaneet2019differ,Yifan:DSS:2019a,handrwr2020} provide a more promising solution for 3D point cloud generation. Similar to unsupervised methods for other 3D representations, such as triangle meshes~\cite{Liu:Paparazzi:2018,liu2018beyond,KatoUH18,liu2019softras,DBLP:journals/corr/abs-1908-01210}, voxel grids~\cite{YanNIPS2016,DBLP:conf/3dim/GadelhaMW17,TulsianiZEM17,mvcTulsiani18}, and implicit functions~\cite{sitzmann2019srns,DIST2019SDFRcvpr,Jiang2019SDFDiffDRcvpr,prior2019SDFRcvpr,NIPS2019_Shichen}, these methods also leverage various differentiable renderers to learn to generate 3D point clouds using 2D images as supervision. In order to recover the 3D structure, the differentiable renderers project the generated 3D point clouds onto a 2D image plane with~\cite{InsafutdinovD18,lin2018learning,Navaneet2019,navaneet2019differ,Yifan:DSS:2019a} or without~\cite{handrwr2020} rendering to compare with the 2D supervision to obtain the per-pixel difference in training, such as density~\cite{InsafutdinovD18,lin2018learning,Navaneet2019,handrwr2020,Yifan:DSS:2019a} or color~\cite{InsafutdinovD18,navaneet2019differ,Yifan:DSS:2019a} error. But these methods are still struggling to recover detailed 3D structures, especially for fine structures like thin tubes or planes.


To resolve this issue, we introduce a novel perspective for unsupervised learning of 3D point cloud generation with fine structure. Different from the current methods, which transform the generated 3D point clouds onto a 2D image plane to compare with regular pixel supervision, we discretize the area covered by the silhouette into discrete and irregular 2D points to compare with the 2D projections of the generated 3D point clouds. Without using the per-pixel difference obtained by various differentiable renderers, we cast the learning of 3D point cloud generation from silhouette images as a 2D projection matching problem. Specifically, rather than using an entire 2D silhouette image as a regular pixel supervision, we discretize the silhouette by randomly sampling a 2D point set within it, which we regard as an irregular point supervision. Then, we push the neural network to generate 3D point clouds whose 2D projections on this silhouette image match the irregular point supervision. One advantage of the irregular point supervision is that it can still capture detailed structure information using sampled points even if fine structures are represented by only few pixels in the silhouette, which is hard for other differentiable renderers to leverage. Our irregular point supervision is robust to different sampling methods and the number of sampled points. Our outperforming results under widely used benchmarks show that our method can recover fine 3D structures from 2D silhouette images at different resolutions. Our contributions are as follows:

\begin{enumerate}[i)]
\item We introduce a method to enable the unsupervised learning of 3D point cloud generation with fine structures by 2D projection matching. Instead of using the per-pixel difference, we introduce irregular point supervision which is sampled from GT silhouettes.
\item We justify the feasibility of the unsupervised learning of 3D point cloud generation using irregular 2D points rather than widely used regular 2D images. This helps to provide more detailed information for fine 3D structures which are represented by only few pixels.
\item We demonstrate that our method can significantly improve the state-of-the-art accuracy in 3D point cloud generation applications by recovering finer structures under various benchmarks.
\end{enumerate}

\section{Related Work}
Deep learning-based 3D shape understanding has achieved very promising results in different tasks~\cite{Park_2019_CVPR,MeschederNetworks,mildenhall2020nerf,wenxin_2020_CVPR,seqxy2seqzeccv2020paper,Groueix_2018_CVPR,Tretschk2020PatchNets,bednarik2020,tancik2020fourfeat,Zhizhong2018VIP,Badki_2020_CVPR,zhizhongiccv2021completing,Zhizhong2021icml,MAPVAE19,p2seq18,hutaoaaai2020,3DViewGraph19,Han2019ShapeCaptionerGCacmmm,wenxin_2021a_CVPR,wenxin_2021b_CVPR,9187572,9318534,Zhizhong2016,Zhizhong2016b,ZhizhongSketch2020,HanCyber17a,HanTIP18,Hu2019Render4CompletionSM,Zhizhong2018seq,3D2SeqViews19,parts4features19,l2g2019,Zhizhong2019seq,DBLP:conf/nips/SitzmannCTSW20,DBLP:journals/corr/abs-2106-03452,DBLP:journals/corr/abs-2105-02788,DBLP:journals/corr/abs-2101-10994,DBLP:journals/corr/abs-2104-10078,DBLP:journals/corr/abs-2106-02634,DBLP:journals/corr/abs-2012-06434,DBLP:journals/corr/abs-2106-05187}. Without 3D supervision, current unsupervised structure learning methods leveraged various differentiable renderers for different 3D raw representations. Differentiable renderers first render a reconstructed 3D shape into 2D silhouette or RGB images, and then, calculate the error between the rendered and GT images to train the neural networks.

\noindent\textbf{Differentiable Renderers for Voxel Grids. }By casting perspective rays through voxel grids, some differentiable renderers rendered images using the maximum occupancy values along each ray~\cite{YanNIPS2016} or the derived ray collision probabilities~\cite{TulsianiZEM17}. While other differentiable renderers employ orthogonal projection using simple projection function~\cite{DBLP:conf/3dim/GadelhaMW17}. These methods work with known camera poses~\cite{YanNIPS2016,TulsianiZEM17,DBLP:conf/3dim/GadelhaMW17}, camera poses estimated from a separate network~\cite{mvcTulsiani18}, or in the presence of viewpoint uncertainties~\cite{Gadelha2019}.

\noindent\textbf{Differentiable Renderers for Triangle Meshes. }OpenDR~\cite{Loper:ECCV:2014} was introduced to approximate gradients with respect to pixel positions in back-propagation. Using hand-crafted gradients, Kato et al.~\cite{KatoUH18} were able to adjust faces on 3D meshes. Similarly, ~\cite{Liu:Paparazzi:2018} and~\cite{liu2018beyond} analytically leveraged computed gradients from images to update face normals along with vertex positions via chain rule. By introducing more advanced rasterization, such as probabilistic rasterization~\cite{liu2019softras} or regarding rasterization as interpolation of local mesh properties~\cite{DBLP:journals/corr/abs-1908-01210}, the pixel value error compared to GT 2D images is used to update mesh reconstruction.

\noindent\textbf{Differentiable Renderers for Implicit Functions. }Implicit functions can represent 3D shapes using occupied voxels or signed distance functions in high resolution, which makes them very popular for deep learning models~\cite{pifuSHNMKL19,xu2019disn,MeschederNetworks,chen2018implicit_decoder,Oechsle_2019_CVPR,Park_2019_CVPR,DBLP:journals/corr/abs-1901-06802,Jiang2019SDFDiffDRcvpr,seqxy2seqzeccv2020paper,takikawa2021nglod,azinovic2021neural}. To reduce the computational cost on sampling points for implicit surface learning, Vincent et al.~\cite{sitzmann2019srns} learned a mapping from world coordinates to a feature representation of local scene properties. Similar to ray marching, various differentiable renderers~\cite{DIST2019SDFRcvpr,Jiang2019SDFDiffDRcvpr,prior2019SDFRcvpr} were proposed to render signed distance functions into images. In addition, ray-based field probing~\cite{shichenNIPS} or aggregating detection points on rays~\cite{DBLP:journals/cgf/WuS20} was leveraged to mine supervision for 3D occupancy fields. With implicit differentiation, Niemeyer et al.~\cite{Volumetric2019SDFRcvpr} analytically derived in a differentiable rendering formulation for implicit shape and texture representations. Moreover, the naturally differentiable volume rendering was also employed to render a learned implicit radiance fields for view synthesis~\cite{mildenhall2020nerf}.

\noindent\textbf{Differentiable Renderers for Point Clouds. }Compactness is an advantage of 3D point clouds, which, however, brings an issue of sparseness among 2D projections of points in rendering. This issue makes it more difficult for differential renderers to directly compare the images rendered from these 2D projections with the ground truth images.

To resolve this issue, different renderers mainly employed either dense 3D points~\cite{lin2018learning} or various rendering approaches~\cite{InsafutdinovD18,Navaneet2019,navaneet2019differ,Yifan:DSS:2019a,Lassner2021,Kolos2020,Grigorev2021,Zhenbo2020,Aliev2020} based on rasterization. Specifically, Lin et al.~\cite{lin2018learning} proposed a pseudo-renderer to render dense points by modeling the visibility using pooling. However, it is significantly affected by the number of points. Instead, rendering based methods~\cite{InsafutdinovD18,Navaneet2019,navaneet2019differ,Yifan:DSS:2019a} rasterized point clouds using surface splatting~\cite{Yifan:DSS:2019a}, Gaussian functions in 3D space~\cite{InsafutdinovD18} or on 2D images~\cite{Navaneet2019,navaneet2019differ}. CapNet~\cite{Navaneet2019} also leveraged a loss to match rendered pixels and GT pixels, however, the rendered pixels are interpolated from multiple 3D point locations, which makes the loss not effective to reveal accurate fine structures by adjusting each 3D point location. Without pixel-wise interpolation, visibility handling, or shading in rendering, DRWR~\cite{handrwr2020} introduced a loss function to directly infer losses for each 3D point from pixel values and 2D projection relationship.

All these methods leveraged per-pixel difference to calculate the gradient in training. This makes it hard to fully recover fine 3D structures. To resolve this issue, we directly match the 2D projections of the generated 3D point clouds to 2D points randomly sampled from the GT silhouette.

\noindent\textbf{Methods without Differentiable Rendering. }Some earlier methods~\cite{Dolphins6165306,Tulsiani7482798,abstractionTulsiani17} did not leverage the rendering strategy to infer 3D structures from 2D images. However, they require strong priors, such as 3D templates~\cite{Dolphins6165306,Tulsiani7482798} or primitives~\cite{abstractionTulsiani17}, and the guidance of 2D and 3D key point correspondences obtained by manual annotation~\cite{Dolphins6165306} or automatic methods~\cite{Tulsiani7482798}, which also makes our method much different.

\section{Method}
\noindent\textbf{Problem Statement. }We aim to learn to generate a 3D point cloud $\bm{M}$ formed by $J$ points $\bm{p}_j$ only using $I$ ground truth silhouette images $\bm{v}_i$, without knowing 3D ground truth $\bm{G}$, where $j\in[1,J]$ and $i\in[1,I]$.

Supervised methods~\cite{Fan_2017_CVPR,cvprpoint2017,nipspoint17,p2seq18,MAPVAE19} can directly train neural networks by minimizing the Chamfer distance (CD) between the generated 3D points $\bm{M}$ and the 3D ground truth $\bm{G}$. Without $\bm{G}$, previous unsupervised methods leveraged the per-pixel difference from the error~\cite{InsafutdinovD18,lin2018learning,Navaneet2019,Yifan:DSS:2019a,navaneet2019differ} between an image $\bm{v}'_i$ rendered with $\bm{M}$ from the the $i$-th view angle and a ground truth image $\bm{v}_i$ or from a loss~\cite{handrwr2020} evaluated on $\bm{v}_i$.

\noindent\textbf{Overview. }Different from the previous differentiable rendering-based methods, we do not leverage the per-pixel difference. Instead, we first discretize the ground truth silhouette images $\bm{v}_i$ by randomly sampling $K$ 2D points $\bm{g}^i_k$, and $k\in[1,K]$ within the silhouette, which we regard as an irregular point supervision. Then, we directly push the 2D projections $\{\bm{q}^i_j\}$ of the generated 3D point cloud $\bm{M}$ on the silhouette image $\bm{v}_i$ to match with the irregular point supervision $\{\bm{g}^i_k\}$. To leverage the supervision from all $I$ view angles, we conduct this 2D projection matching procedure on each of the views.

\begin{figure}[tb]
  \centering
   \includegraphics[width=\linewidth]{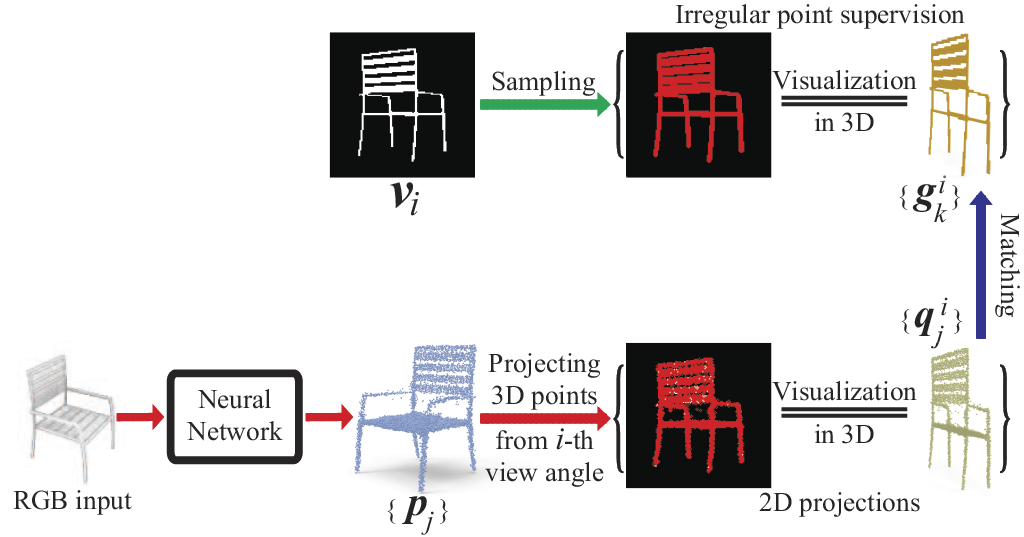}
  %
  %
\caption{\label{fig:frameworks} Overview of our method. We train the neural network to learn to generate 3D point clouds $\{\bm{p}_j\}$ by pushing its projection $\{\bm{q}^i_j\}$ to match with irregular point supervision $\{\bm{g}^i_k\}$. We obtain $\{\bm{g}^i_k\}$ by randomly sampling 2D points to cover a silhouette on ground truth silhouette image.}
\end{figure}

We demonstrate our method in Fig.~\ref{fig:frameworks}. Our purpose is to train a neural network to learn a mapping from an RGB input to a 3D point cloud $\bm{M}$ formed by a set of 3D points $\{\bm{p}_j\}$. To achieve this, we project the generated 3D points $\{\bm{p}_j\}$ onto one silhouette image $\bm{v}_i$ to get 2D projections $\{\bm{q}^i_j\}$ of $\bm{M}$. Then, we push projections $\{\bm{q}^i_j\}$ to match the irregular point supervision $\{\bm{g}^i_k\}$, which is randomly sampled within the silhouette on $\bm{v}_i$.

\noindent\textbf{Advantages. }By removing the requirement of the per-pixel difference, our method has several advantages over current differentibale rendering based methods.

Our method is simpler, since we do not require any complex and time consuming rendering procedures~\cite{InsafutdinovD18,lin2018learning,Navaneet2019,navaneet2019differ,Yifan:DSS:2019a} such as surface interpolation, visibility handling, and shading.

Although DRWR~\cite{handrwr2020} does not require rendering either, it is hard to force the 2D projections to uniformly cover the silhouettes with fine structures by merely repulsing pairwise projections within the silhouettes. In contrast, our method is more effective by directly representing the fine structures using irregular point supervision.

Rendering based methods~\cite{InsafutdinovD18,lin2018learning,Navaneet2019,navaneet2019differ,Yifan:DSS:2019a} are sensitive to the resolution of the ground truth silhouette images since they need to compare in a pixel-by-pixel manner. In contrast, we discretize the silhouette into irregular point supervision that is independent of image resolution. This enables us to formulate a loss based on 2D point matching, which we find to be more effective in our experiments.

\begin{figure}[tb]
  \centering
   \includegraphics[width=\linewidth]{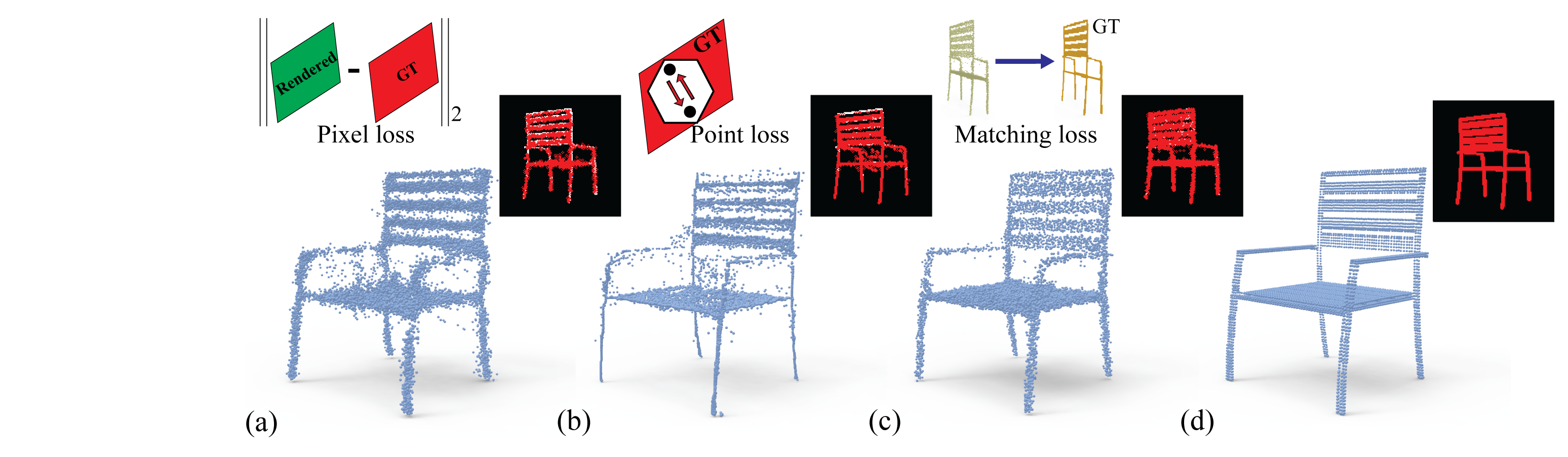}
  %
  %
\caption{\label{fig:rendercom} Overfitting experiment using different loss functions. Methods with rendering (a) and without rendering (b) leverage per-pixel difference to the ground truth 2D supervision (d) to infer 3D structures, while our method (c) provides a different perspective using 2D projection matching, where each of $16000$ 2D projections is shown in red on silhouette images.}
\end{figure}

We highlight our advantages over current differentiable renderers using an overfitting demonstration in Fig.~\ref{fig:rendercom}. Current differentiable renderers depend on the per-pixel difference to evaluate how well the 2D projections cover the silhouette by using a rendering based pixel loss~\cite{InsafutdinovD18,lin2018learning,Navaneet2019,navaneet2019differ,Yifan:DSS:2019a} in Fig.~\ref{fig:rendercom} (a) or a rendering free based point loss~\cite{handrwr2020} in Fig.~\ref{fig:rendercom} (b). The projections shown on the top right in each subfigure demonstrate that current differentiable renderers cannot fully cover the fine structures on silhouette images, compared to the ground truth in Fig.~\ref{fig:rendercom} (d), which significantly affects the 3D point cloud generation.

\begin{figure}[tb]
  \centering
   \includegraphics[width=\linewidth]{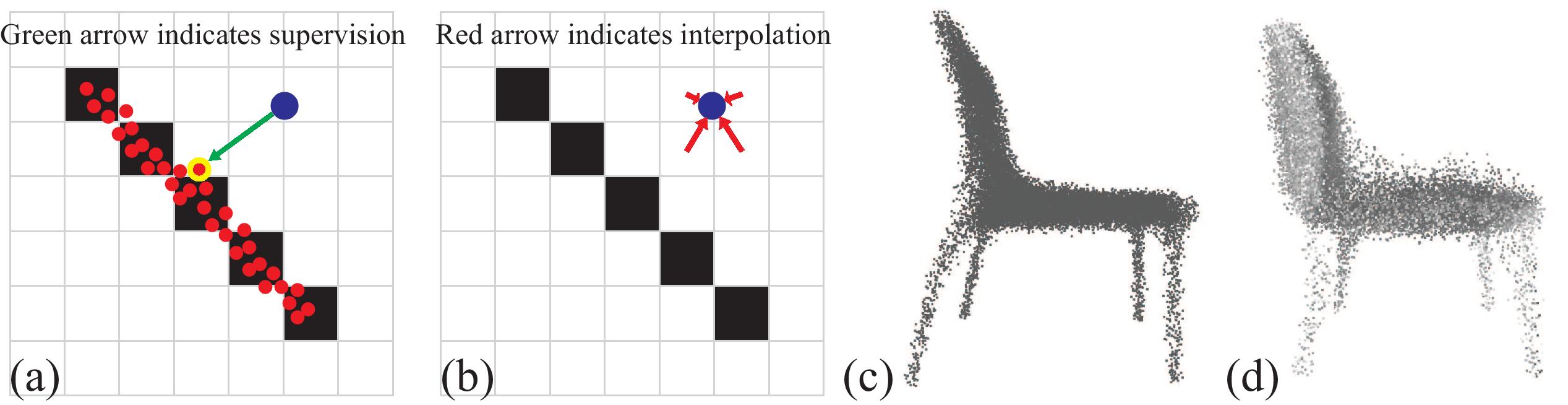}
  %
  %
\caption{\label{fig:difference} Our dense sampled points (red) in (a) are more effective supervision for projection (blue) to reveal fine structures (black) than per-pixel difference with interpolated pixel values in (b). The length of gradient for each point using our irregular point supervision or per-pixel difference is shown as color in (c) and (d).}
\end{figure}

Our method resolves this issue by directly matching 2D projections with the explicitly irregular point supervision. As shown in Fig.~\ref{fig:difference} (a), our irregular point supervision can provide more specific and accurate supervision as dense sampled points in the fine structure represented by merely few pixels, which is more effective than minimizing per-pixel difference using interpolated pixel values shown in Fig.~\ref{fig:difference} (b). During optimization, our much smaller length (darker color) of gradient for each point  in Fig.~\ref{fig:difference} (c) than the gradient with per-pixel difference in Fig.~\ref{fig:difference} (d) demonstrates that our irregular point supervision can provide a more clear target for each point, which is much easier to achieve.

\noindent\textbf{Projecting 3D Points. }We generate a 3D point cloud $\bm{M}$ in an object centered coordinate system. We leverage perspective projection to project 3D points $\{\bm{p}_j\}$ on $\bm{M}$ as 2D projections $\{\bm{q}^i_j\}$ on each silhouette image $\bm{v}_i$ from the $i$-th view angle. We denote $\bm{C}_i$ as both extrinsic and intrinsic camera parameters of the $i$-th camera pose, so we can perform the perspective projection as below,

\begin{equation}
\label{eq:transform}
[\bm{q}_j^i \quad 1]^T\sim\bm{C}_i[\bm{p}_j \quad 1]^T.
\end{equation}

\noindent\textbf{Irregular Point Supervision. }We aim to provide the training supervision by discretizing a silhouette on image $\bm{v}_i$ through sampling the silhouette into irregular point supervision $\{\bm{q}^i_j\}$, which fully covers the silhouette. According to our preliminary results, we found that the consistency of points sampled on different silhouette images affects the performance, since the 2D projections on different silhouette images are from the same generated 3D point cloud. Although this issue can be alleviated by sampling very dense points within each silhouette with many different sampling methods, it would make the loss calculation more costly.

To improve the consistency of points sampled within the silhouette from different view angles, we introduce a \textit{Structure Adaptive Sampling} (SAS) method to sample points within each silhouette. SAS first determines a ratio $r$ to indicate how much area each one of $K$ sampled points can cover within a silhouette. We calculate the area $A$ of the whole silhouette by counting the number of pixels with a value of 1. So, the ratio $r$ equals to $A/K$. Then, we calculate the sampling step $s$ as $\sqrt{r}$, which is the edge length of the area that each sampled point covers. Finally, we start the sampler from the coordinate of $(0,0)$ with a stride of $s$, and sample one point at each step if its interpolated pixel value is larger than 0.5.

\noindent\textbf{Loss Function. }Our method casts the learning of 3D point cloud generation into a 2D projection matching problem. Therefore, we push the neural network to generate a 3D point cloud $\bm{M}$ whose projections $\{\bm{q}^i_j\}$ on the $i$-th view can match the corresponding irregular point supervision $\{\bm{g}^i_k\}$, where we conduct the matching on each one of the $I$ views.

On the $i$-th view, we leverage the CD to evaluate the distance between $\{\bm{q}^i_j\}$ and $\{\bm{g}^i_k\}$ as follows,

\begin{equation}
\label{eq:cd}
\begin{aligned}
d(\{\bm{q}^i_j\},\{\bm{g}^i_k\})=&\frac{1}{J}\sum_{q\in\{\bm{q}^i_j\}}\min_{g'\in\{\bm{g}^i_k\}}||q-g'||_2^2\\
&+\frac{1}{K}\sum_{g\in\{\bm{g}^i_k\}}\min_{q'\in\{\bm{q}^i_j\}}||g-q'||_2^2,
\end{aligned}
\end{equation}

\noindent Our preliminary results show that the Earth Mover Distance (EMD) can also get comparable results to CD when evaluating the matching distance. But EMD requires the number of the 2D projections and the number of irregular point supervision to be the same, which may not be necessary to sample many points within each silhouette.

We conduct the 2D projection matching on all $I$ views by minimizing a loss function $L$ below,

\begin{equation}
\label{eq:L}
L=\sum^{i=I}_{i=1}  d(\{\bm{q}^i_j\},\{\bm{g}^i_k\}).
\end{equation}

\section{Experiments, Analysis and Discussion}
\subsection{Experimental Details}
\noindent\textbf{Dataset. }For the fair comparison with the previous methods~\cite{InsafutdinovD18,lin2018learning,Navaneet2019,navaneet2019differ,Yifan:DSS:2019a}, we evaluate our method using the same three categories from ShapeNet~\cite{ChangFGHHLSSSSX15}, including chairs, cars, and airplanes. We also keep the same train/test splitting as in~\cite{TulsianiZEM17,InsafutdinovD18,handrwr2020}, and we employ the benchmark released by~\cite{InsafutdinovD18}, which is formed by the rendered images and ground truth point clouds. For each 3D shape, there are $I=5$ rendered views at three different resolutions including $32^2$, $64^2$, and $128^2$, while all of them correspond to the same ground truth point clouds. Note that the ground truth point clouds have different numbers of points.

\begin{figure}[]
  \centering
   \includegraphics[width=\linewidth]{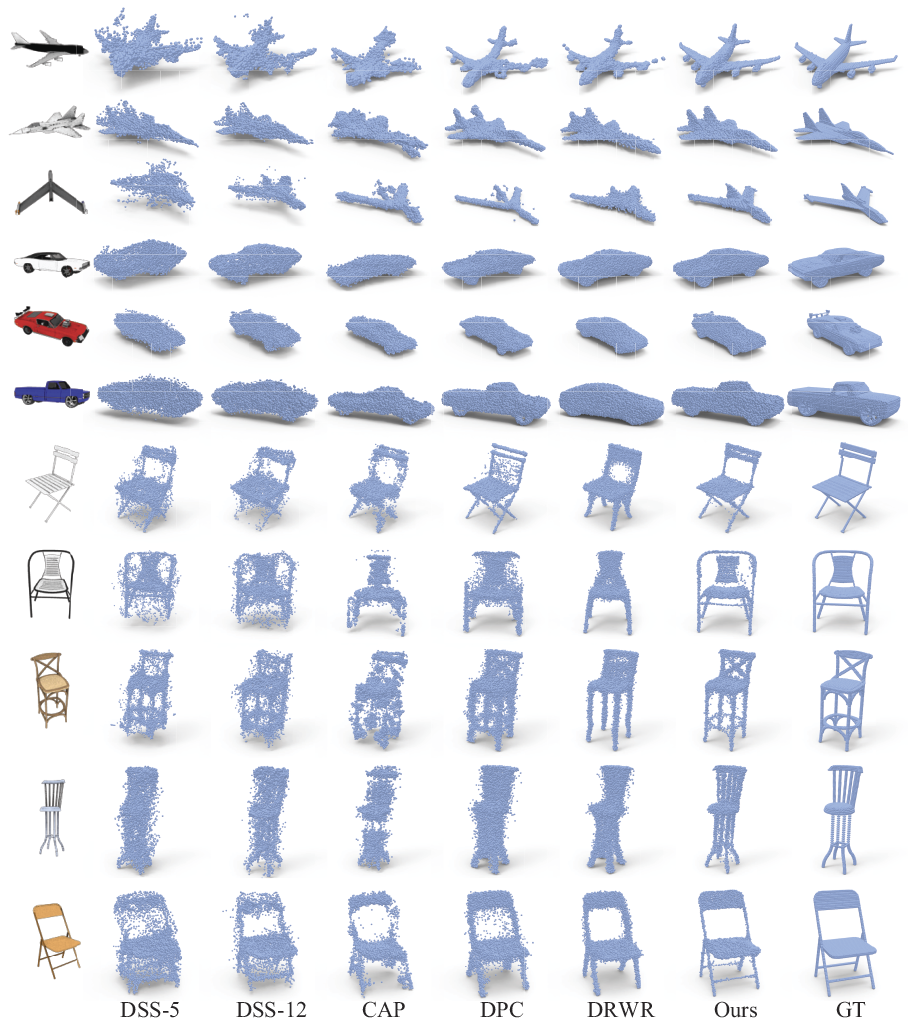}
  %
  %
\caption{\label{fig:Comparison} Visual comparison ($16000$ points) with the state-of-the-art methods using synthetic images.}
\vspace{-0.28in}
\end{figure}

\noindent\textbf{Metric. }We evaluate our results using the CD between the predicted and the ground truth 3D point clouds. To compare with differentiable renderers for different 3D representations such as meshes or voxel grids, we also use volumetric IoU at a resolution of $32^3$ to conduct fair comparisons. For better readability, we multiply all CD or IoU values reported in our experiments by 100.

\noindent\textbf{Setup. }We realize that the neural network structure may affect the performance. Therefore, we employ the same neural network which was also employed by the previous differentiable renderers~\cite{InsafutdinovD18,handrwr2020} in evaluation. We will elaborate on the network structure in our supplemental material.

For training, we use image pairs from the same 3D shape to leverage the supervision from multiple views as the previous methods~\cite{InsafutdinovD18,handrwr2020}. For each pair of the $I=5$ views, we use one RGB image as the input to generate a 3D point cloud $\bm{M}$ from the network, and then project $\bm{M}$ onto the other image to push the 2D projections $\{\bm{q}^i_j\}$ to match with the corresponding irregular point supervision $\{\bm{g}^i_k\}$.

We project the generated 3D point cloud $\bm{M}$ using the known camera pose although we can also leverage another network to estimate the camera pose from the input RGB image as~\cite{InsafutdinovD18,mvcTulsiani18}. This is because camera pose estimation itself is another challenging problem which may affect the evaluation of 3D structure learning performance.

In addition, we train our network using all the three kinds of image resolutions provided in the benchmark~\cite{InsafutdinovD18} respectively, and accordingly generate point clouds at three different resolutions $J\in[2000,8000,16000]$.

We train our network using the Adam optimizer with a learning rate of $0.0001$. Each batch contains 16 rendered images which are equally from 4 shapes, where we iterate over $5\times10^6$ batches in each experiment. We initially sample $K=5000$ points within each silhouette from each view angle to form the irregular point supervision.

\subsection{Single Image Reconstruction}
We first evaluate our method in single image reconstruction. We train the neural network using the rendered images from the benchmark, while testing the network using the synthetic image or real image respectively.

\noindent\textbf{Synthetic Image Testing. }We compare our method with the latest methods leveraging the per-pixel difference. These methods include Differentiable Surface Splatting (DSS)~\cite{Yifan:DSS:2019a}, Differentiable Ray Consistency (DRC)~\cite{TulsianiZEM17}, Efficient Point Cloud Generation (EPCG)~\cite{lin2018learning}, Continuous Approximation Projection (CAP)~\cite{Navaneet2019}, Differentiable Point Clouds (DPC)~\cite{InsafutdinovD18}, and Differentiable Renderer without Rendering (DRWR)~\cite{handrwr2020}. DRC is voxel-based, and it is only available for voxel grids at a resolution of $32^3$ because of the cubic complexity of voxel grids. The other four renderers are point cloud-based.

We first report the numerical comparison in terms of CD in Table~\ref{table:t10}. Our results are the best under all classes at all three resolutions. Our method shows significant improvement over voxel-based differentiable renderers including DRC and the voxel-based counterpart ``DPC-V'' of DPC~\cite{InsafutdinovD18}. Moreover, our method also reconstructs much more accurate point clouds compared to differentiable point renderers using per-pixel difference, such as CAP~\cite{Navaneet2019}, DPC~\cite{InsafutdinovD18}, and EPCG~\cite{lin2018learning}. By leveraging 2D projection matching, our method is able to reconstruct 3D structures with high accuracy, especially for fine structures like thin planes and tubes. To demonstrate this, we conduct visual comparison with DSS~\cite{Yifan:DSS:2019a}, CAP~\cite{Navaneet2019}, DPC~\cite{InsafutdinovD18}, DRWR~\cite{handrwr2020} in Fig.~\ref{fig:Comparison} using the reconstructed shapes in the test set, where DSS reconstructed point clouds with 16000 points from 5 views (``DSS-5'') which are the same ones used by other methods or 12 views (``DSS-12''). The comparison shows that our method can reveal more 3D structure details, such as the complex structures on the chairs, cars and airplanes. Please see more point cloud reconstructions in our supplemental material.

\begin{table*}[tb]
\centering
\caption{Numerical comparison with differentiable renderers in terms of CD.}  
\resizebox{\linewidth}{!}{
    \begin{tabular}{c|c|c|c|c|c|c|c|c|c|c|c|c|c|c}  
     \hline
        & \multicolumn{6}{|c|}{Image-$32^2$, Shape-2000}  & \multicolumn{4}{|c|}{Image-$64^2$, Shape-8000} & \multicolumn{4}{|c}{Image-$128^2$, Shape-16000} \\
       \hline
        & DRC & CAP & DPC-V & DPC &DRWR&Ours& DPC-V & DPC& DRWR &Ours& EPCG & DPC&DRWR&Ours \\  
     \hline
     Plane &8.35&6.34&5.57&4.52&4.01&\textbf{3.37}&4.94&3.50&3.18&\textbf{2.94}&4.03&2.84&2.66&\textbf{2.08}\\
     Car&4.35&6.03&3.88&4.22&3.81&\textbf{3.50}&3.41&2.98&2.89&\textbf{2.81}&3.69&2.42&2.40&\textbf{2.25}\\
     Chair&8.01&6.11&5.57&5.10&4.66&\textbf{4.16}&4.80&4.15&4.02&\textbf{3.94}&5.62&3.62&3.49&\textbf{3.10}\\
     \hline
     Mean&6.90&6.16&5.01&4.61&4.16&\textbf{3.68}&4.39&3.55&3.36&\textbf{3.23}&4.45&2.96&2.85&\textbf{2.48}\\
     \hline
   \end{tabular}}
   \label{table:t10}
\end{table*}

Moreover, we also compare our method with the latest supervised 3D point cloud generation method called NOX~\cite{NIPS2019_Srinath}. We report our results using the same evaluation code and setting released by NOX~\cite{NIPS2019_Srinath} in Table~\ref{table:NOX}. Specifically, we scale the point clouds reconstructed from input images at a resolution of $64^2$ in Table~\ref{table:t10}, so that the diagonal of the bounding box of each reconstructed point cloud is one. We also resample the ground truth point cloud in the benchmark released by~\cite{InsafutdinovD18} to 8000 points which keeps the number of points the same as NOX~\cite{NIPS2019_Srinath}.

\begin{table}[h]
\centering
\caption{Comparison with supervised point generation method.}  
\resizebox{0.7\linewidth}{!}{
    \begin{tabular}{c|c|c|c}  
     \hline
          CD& Cars & Airplanes & Chairs\\   
     \hline
       NOX& 0.1569 & 0.1855 & 0.3803 \\
       Ours & \textbf{0.0421} & \textbf{0.0492} & \textbf{0.0529} \\
     \hline
   \end{tabular}}
   \label{table:NOX}
\end{table}

\begin{figure}[]
  \centering
   \includegraphics[width=\linewidth]{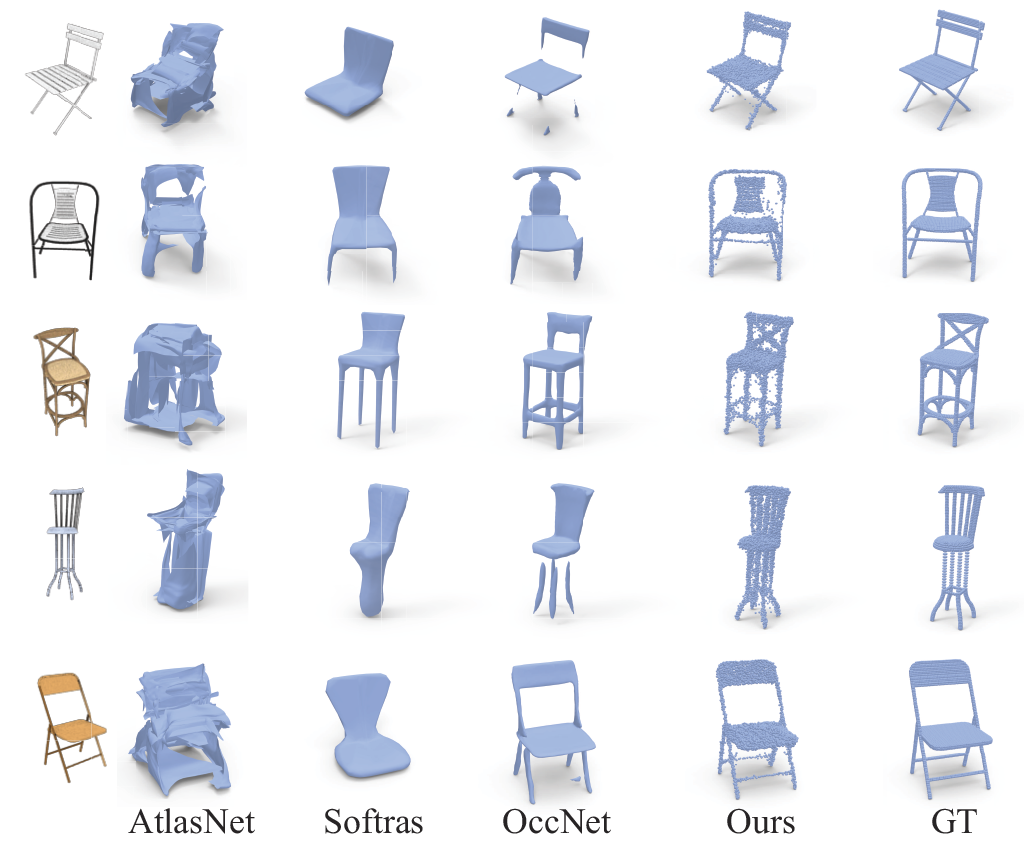}
  %
  %
\caption{\label{fig:Comparison1} Visual comparison with the state-of-the-art supervised and unsupervised methods for different 3D representations.}
\vspace{-0.19in}
\end{figure}

Finally, we conduct a numerical comparison in terms of IoU with other supervised or unsupervised 3D shape generation methods for different 3D shape representations including triangle meshes, voxel grids, point clouds and implicit functions. The compared differentiable renderers include Perspective Transform Nets (PTN)~\cite{YanNIPS2016}, Neural Mesh Renderer (NMR)~\cite{KatoUH18}, SoftRasterizer (SoftRas)~\cite{liu2019softras}, Interpolation-based Differentiable Renderer (DIB-R)~\cite{DBLP:journals/corr/abs-1908-01210}, Implicit Surface renderer (IMRender)~\cite{NIPS2019_Shichen}, Implicit Function renderer (IMFun)~\cite{DBLP:journals/cgf/WuS20}, and SDFDiff~\cite{Jiang2019SDFDiffDRcvpr}. The first method is voxel-based, the following two methods are mesh-based, while the last three are implicit function based. We report the results of NMR, SoftR and DIB-R from~\cite{DBLP:journals/corr/abs-1908-01210}, and the rest from the original papers. The supervised methods in comparison include DISN~\cite{xu2019disn}, OccNet~\cite{MeschederNetworks}, IMNET~\cite{chen2018implicit_decoder}, 3DN~\cite{wang20193dn}, Pix2Mesh~\cite{WangZLFLJ18}, R2N2~\cite{ChoyXGCS16}, and AtlasNet~\cite{Groueix_2018_CVPR}. To report our IoU results, we voxelize the point clouds predicted from images at a resolution of $128^2$ in Table~\ref{table:t10} into voxel grids at a resolution $32^3$ to compare to the same ground truth as other methods. Our outperforming results in Table~\ref{table:VOXEL} demonstrate our advantage in fine structure generation for 3D shapes. Note that although SDFDiff performs a little bit better under the Chair class, it uses RGB images as the supervision signal and requires to know the illumination and surface reflectance model, while our method does not require any such information. Fig.~\ref{fig:Comparison1} demonstrates that our method can learn more complex structures on chairs than methods for meshes (``AtlasNet'',``Softras'') and implicit functions (``OccNet''), where we produced their results using the trained models released from their papers~\cite{Groueix_2018_CVPR,liu2019softras,MeschederNetworks}.

\begin{table*}[ht]
\centering
\caption{Numerical comparison with supervised and unsupervised 3D shape generation methods in terms of IoU.}  
\resizebox{\linewidth}{!}{
    \begin{tabular}{c|c|c|c|c|c|c|c|c|c||c|c|c|c|c|c|c|c}  
     \hline
     &\multicolumn{9}{c||}{Unsupervised differentiable renderers} &\multicolumn{7}{|c|}{Supervised structure learning methods}& \\
     \hline
          & PTN & NMR & SoftRas&DIB-R& IMRender&IMFun&SDFDiff&DRWR& Ours&DISN&OccNet&IMNET&3DN&Pix2Mesh&R2N2&AtlasNet&Ours\\ 
     \hline

     Car &71.2&71.3&77.1&78.8&78.2&66.0&80.0&75.3&\textbf{80.2}&74.3&73.7&74.5&59.4&50.1&66.1&22.0&\textbf{80.2}\\
     Plane&55.6&58.5&58.4&57.0&65.1&53.3&68.7&62.2&\textbf{69.9}&57.5&57.1&55.4&54.3&51.5&42.6&39.2&\textbf{69.9}\\
     Chair&44.9&41.4&49.7&52.7&54.8&44.4&\textbf{64.4}&58.1&62.7&54.3&50.1&52.2&34.4&40.2&43.9&25.7&\textbf{62.7}\\
     \hline
     Mean&57.2&57.1&61.7&62.8&66.0&54.6&\textbf{71.0}&65.2&\textbf{71.0}&62.0&60.3&60.7&49.4&47.3&50.9&29.0&\textbf{71.0}\\
     \hline
   \end{tabular}}
   \label{table:VOXEL}
\end{table*}

\begin{figure}[tb]
  \centering
   \includegraphics[width=\linewidth]{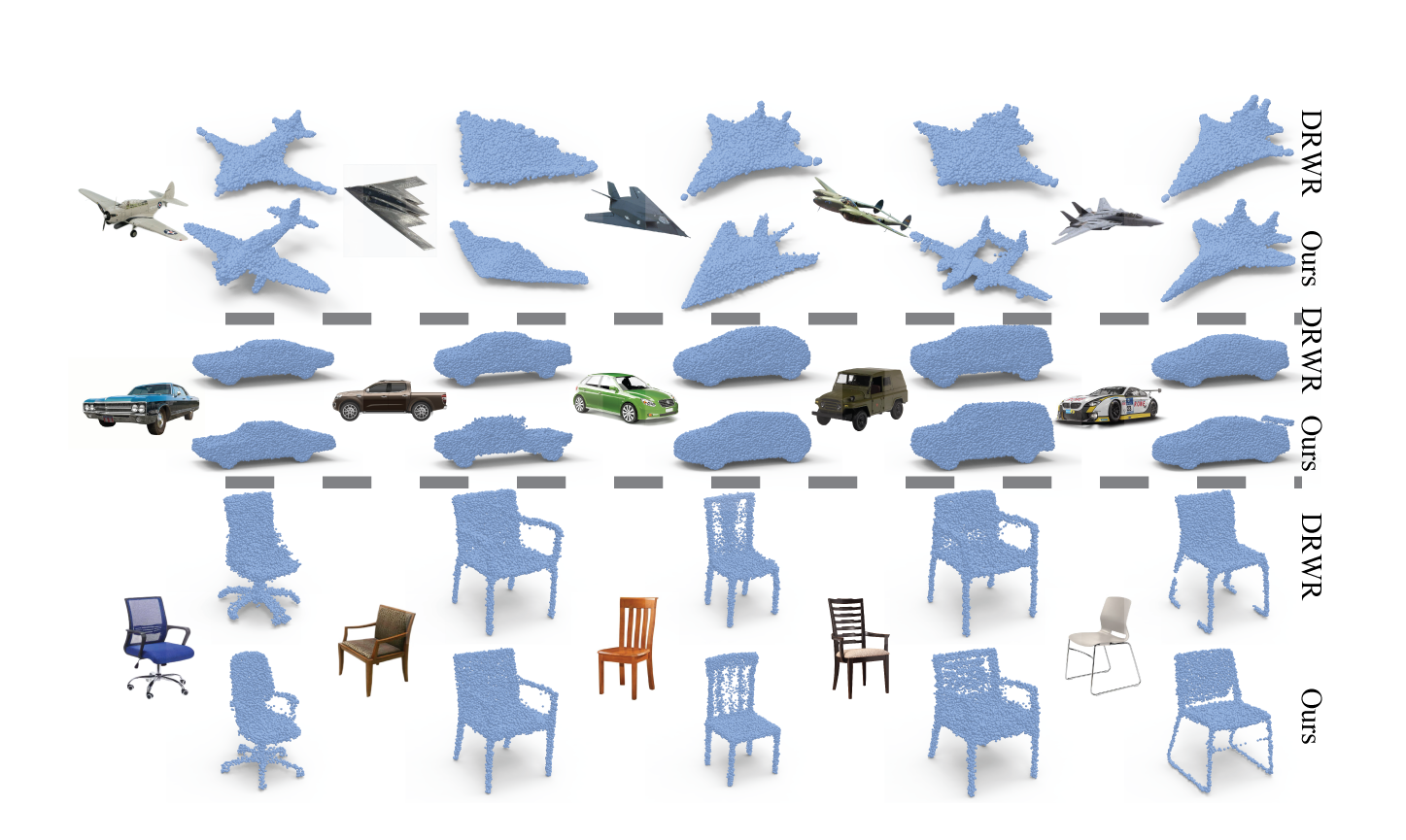}
  %
  %
\caption{\label{fig:RealComp} Visual reconstruction comparison using real images.}
\vspace{-0.28in}
\end{figure}

\noindent\textbf{Real Image Testing. }Next, we evaluate our trained neural network by testing its adaptation to real images containing fine 3D structures. We randomly select some real images from the Internet, and use them as input to generate 3D point clouds at a resolution of 16000 points using the parameters learned in Table~\ref{table:t10}. As a comparison, we also generate the point clouds at the same resolution from the same images using parameters learned by DRWR~\cite{handrwr2020}. Fig.~\ref{fig:RealComp} demonstrates that our method can adapt better to real images and reconstruct high fidelity 3D point clouds with more accurate fine structures than DRWR, such as the thin wings of airplanes and the thin legs of chairs.

\subsection{Novel Shape Generation}
We further evaluate our method in another task of novel shape generation. We conduct this experiment under the Chair class to learn to generate novel point clouds at a resolution of 16000 from $128^2$ images. Specifically, we modify our current encoder-decoder network structure by adding a KL loss on the latent code between the image encoder and the point decoder, which makes it very similar to the structure of Variational Auto Encoder (VAE)~\cite{KingmaW13}. We train this network using a loss function formed by the matching loss in Eq.~\ref{eq:cd} and the KL loss with a balance ratio of $1:5\times10^{-6}$. In this way, we map the latent code of each input RGB image into a 32 dimensional Gaussian space during training so that we can generate a novel 3D point cloud from a randomly sampled latent code in the Gaussian space using the trained point decoder during inference.

\begin{figure}[tb]
  \centering
   \includegraphics[width=\linewidth]{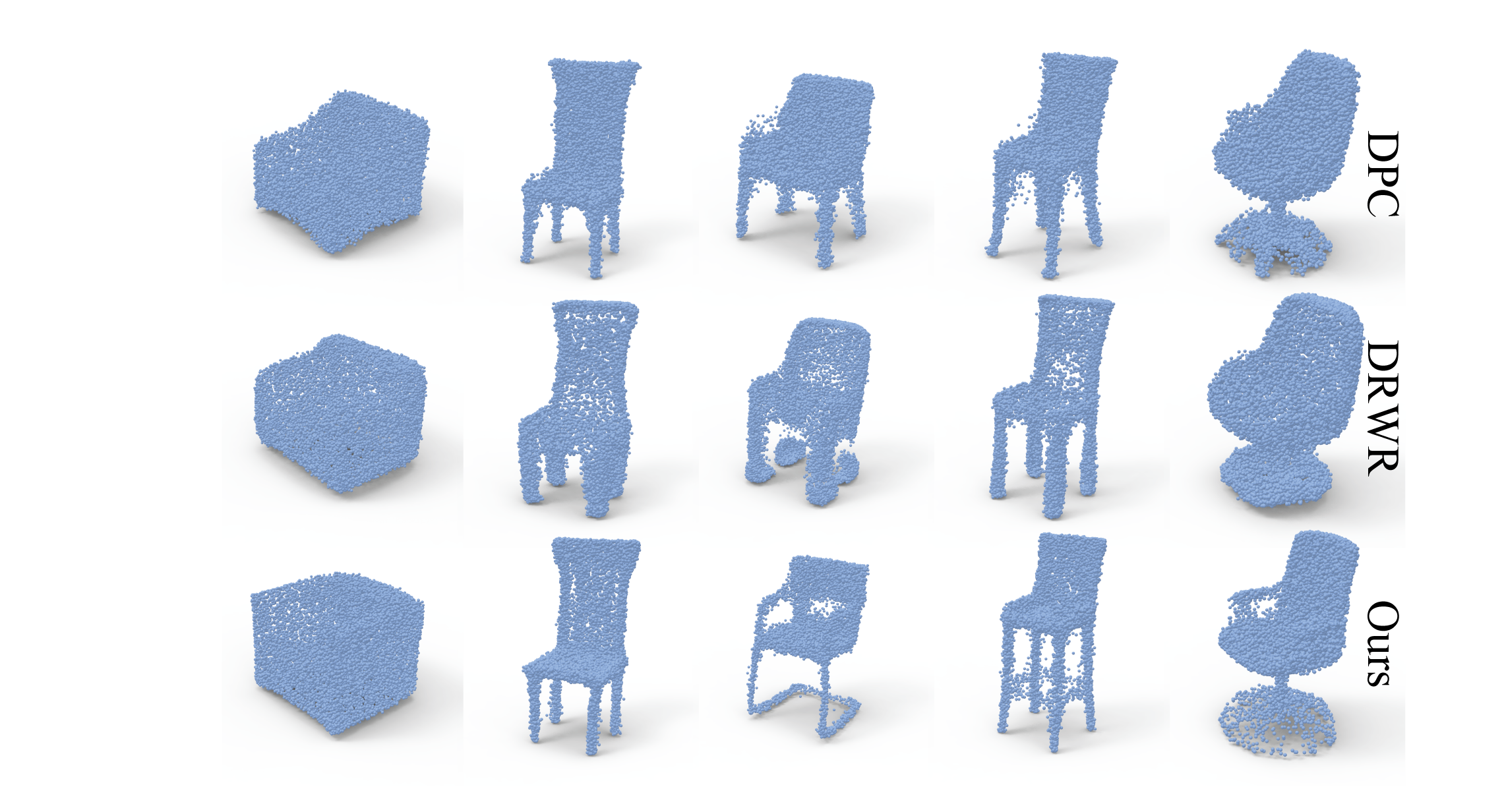}
  %
  %
\caption{\label{fig:NovelComp} Visual comparison in novel shape generation.}
\vspace{-0.28in}
\end{figure}

We compare our method with DPC~\cite{InsafutdinovD18} and DRWR~\cite{handrwr2020} by generating 3D point clouds using the randomly sampled latent codes in Fig.~\ref{fig:NovelComp}, where the three point clouds in each column are generated by the same latent code. The visual comparison demonstrates that our method can help the network to capture more fine 3D structures from the 2D supervision during training, which can be further leveraged to generate more reasonable and plausible 3D point clouds with fine structures, such as thin legs of chairs. Please see more point cloud generation in supplemental material.

\subsection{Analysis and Discussion}
\noindent\textbf{Ablation Studies. }We explore the effectiveness of the two terms in the loss function in Eq.~\ref{eq:cd}. We conduct these experiments by merely using either one of the two terms to learn to generate 3D point clouds at a resolution of 2000 from $32^2$ images under the Chair class. The degenerated results of ``1st'' and ``2nd'' in Table~\ref{table:ablation} show that the one directional distance loss cannot push the 2D projections to fully match with the irregular point supervision. To explore whether we can better resist the inconsistency among the irregular point supervision from different view angles, we adjust the number of nearest neighbors involved in the two terms in Eq.~\ref{eq:cd}. We tried to leverage more nearest neighbors, such as 5 in the first term and 1 in the second term (``NN(5,1)''), 1 in the first term and 5 in the second term (``NN(1,5)''), or 5 in both terms (``NN(5,5)''). However, we do not observe any improvement compared to our original setting (``NN(1,1)'').

\begin{table}[h]
\centering
\caption{Ablation studies in terms of CD.}  
\resizebox{\linewidth}{!}{
    \begin{tabular}{c|c|c|c|c|c|c}  
     \hline
          & 1st & 2nd & NN(5,1) & NN(1,5) & NN(5,5) &NN(1,1)\\   
     \hline
       CD &24.47&8.49&4.25&4.32&4.29&\textbf{4.16}\\
     \hline
   \end{tabular}}
   \label{table:ablation}
\end{table}

\noindent\textbf{Resolution of Irregular Point Supervision. }The resolution of irregular point supervision is also important in training. If the number of points in the irregular point supervision is too small, the inconsistency among different view angles will be enlarged, which significantly affects the inference of fine structures on the 3D point cloud. On the other hand, if the number of points in the irregular point supervision is too large, it may bring redundancy without improving the learning performance, which also increases the computational burden when calculating the loss. To explore the trade-off, we sample different numbers of points within the silhouette to form the irregular point supervision, such that $K=\{1000,3000,5000,7000,9000\}$. Using the irregular point supervision at each resolution, we train a network to learn to generate 3D point clouds at a resolution of 2000 from $32^2$ images under the Chair class. The results in Table~\ref{table:number} show that $K=5000$ achieves the best accuracy and no improvement is observed with a larger $K$. Therefore, we use $K=5000$ to establish irregular point supervision under different shape classes. In addition, the obtained results are also robust to different resolutions, since the results just change a little if the resolution is larger than $K=3000$.

\begin{table}[h]
\centering
\caption{Resolution of irregular point supervision comparison.}  
\resizebox{0.8\linewidth}{!}{
    \begin{tabular}{c|c|c|c|c|c}  
     \hline
         $K$ & 1000 & 3000 & 5000 & 7000 & 9000\\   
     \hline
       CD &4.29&4.20&\textbf{4.16}&4.20&4.19\\
     \hline
   \end{tabular}}
   \label{table:number}
\end{table}

\noindent\textbf{Sampling for Irregular Point Supervision. }The sampling for establishing irregular point supervision also affects the performance because of the inconsistency among irregular point supervision from different view angles. Our structure adaptive sampling (``SAS'') can alleviate the inconsistency to infer more accurate fine structures on 3D point clouds. We produce irregular point supervision with $K=5000$ to generate shapes in 2000 points from $32^2$ images under the Chair class using random sampling (``Rand''), pixel sampling (``Pixel''), pixel and random sampling (``Pix+Ran''), and Poisson-Disk sampling (``Poisson'')~\cite{DBLP:conf/siggraph/Bridson07}.

\begin{figure*}[tb]
  \centering
   \includegraphics[width=0.8\linewidth]{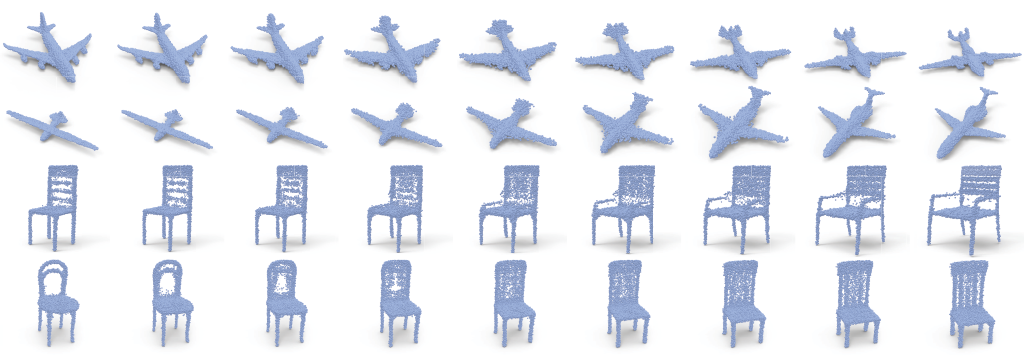}
  %
  %
\caption{\label{fig:InterpolationShapes} Shape interpolation demonstration to visualize the latent code space.}
\vspace{-0.28in}
\end{figure*}

Here, random sampling randomly samples a point on the silhouette image, and leverages a threshold of 0.5 to determine whether keeping this point in irregular point supervision if its interpolated pixel value is larger than 0.5, until the number of sampled points reaches $K=5000$. While pixel sampling just leverages the pixel locations with an interpolated pixel value larger than 0.5. If the number of pixels is smaller than $K=5000$, we will repeat the sampled pixel locations. ``Pix+Ran'' is similar to pixel sampling, but it uses randomly sampled points to replace the repeating procedure. Moreover, we also tried to do random sampling in a dynamic way, which aims to randomly sample $K=5000$ points in each epoch to alleviate the impact of inconsistency using the randomness, as shown by the result of ``Dynamic''.

The comparison in Table~\ref{table:number1} shows that our SAS sampling achieves the best inference performance for 3D fine structures during training due to the produced more consistent irregular point supervision. Random sampling in a dynamic way cannot improve the performance either. Moreover, we found that our method is robust to different sampling methods although SAS sampling could help the neural network to learn the most accurate structure. This is because the results with different sampling methods do not change a lot, and more importantly, all of them are better than the state-of-the-art result of $4.66$ obtained by DRWR in Table~\ref{table:t10}.

\begin{table}[h]
\centering
\caption{Sampling for irregular point supervision.}  
\resizebox{\linewidth}{!}{
    \begin{tabular}{c|c|c|c|c|c|c}  
     \hline
          & Rand & Pixel & Pix+Ran & Poisson & Dynamic& SAS\\   
     \hline
       CD &4.23&4.20&4.22&4.35&4.36&\textbf{4.16}\\
     \hline
   \end{tabular}}
   \label{table:number1}
   \vspace{-0.19in}
\end{table}

\noindent\textbf{Latent Code Visualization. }We visualize the latent space learned in the network which is trained to produce our results of 16000 points in Table~\ref{table:t10}. We randomly select two reconstructed point clouds in the test set, and employ their latent codes to interpolate seven new latent codes between them which are further used to generate seven novel shapes by the trained point decoder. We visualize two pairs of shape interpolation under each one of Airplane and Chair classes in Fig.~\ref{fig:InterpolationShapes}. The smooth transformation from one shape to another shape demonstrates that our method can help the network to learn a meaningful latent space.

\begin{figure}[]
  \centering
   \includegraphics[width=\linewidth]{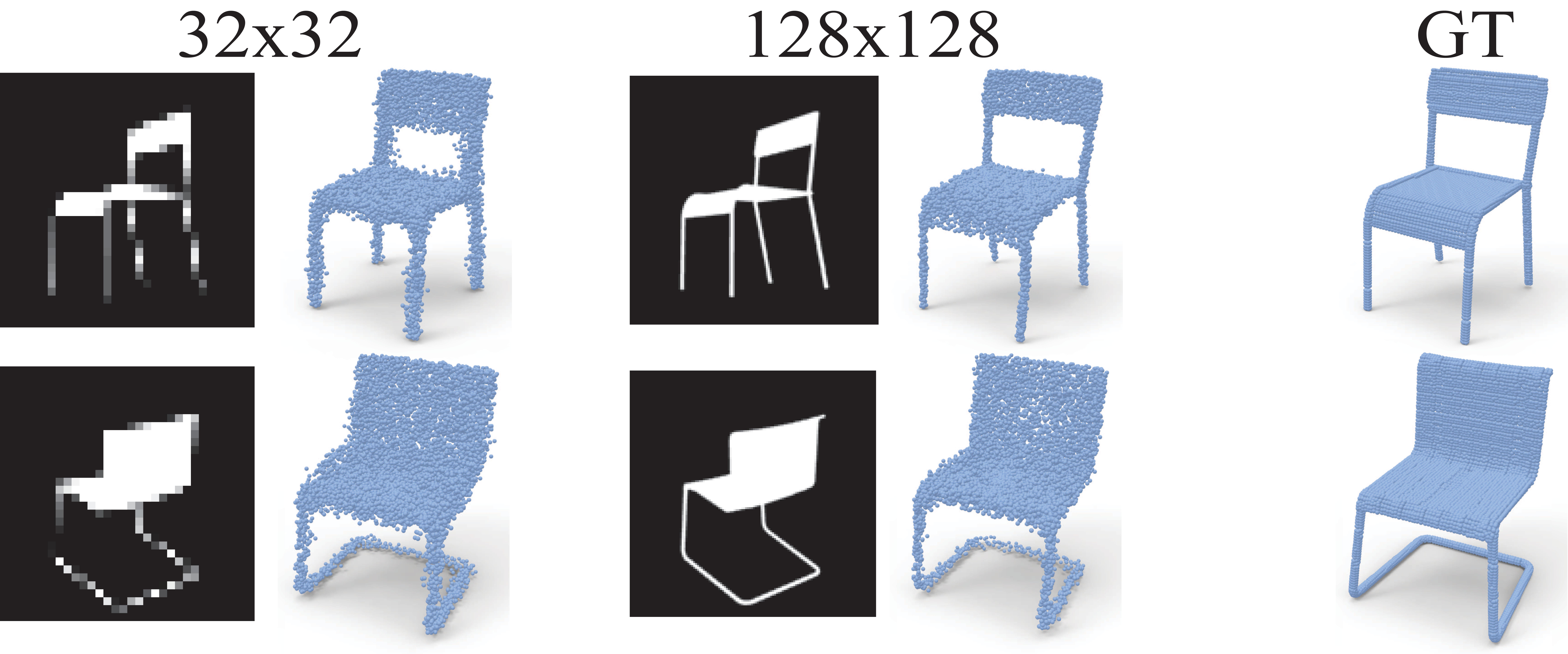}
  %
  %
\caption{\label{fig:Comparison2} Robustness to the 2D supervision resolution.
}
\vspace{-0.28in}
\end{figure}

\noindent\textbf{Robustness to 2D supervision resolution. }Without using the per-pixel difference, our method is more robust to the resolution of 2D supervision. To demonstrate this, we train the neural network under the Chair class to generate 3D point clouds at a resolution of 16000 points from $32^2$ images, rather than $128^2$ images shown in Table~\ref{table:t10}. Due to the degenerated structure information caused by the lower resolution of 2D supervision, it is reasonable to obtain degenerated results. Therefore, we compare our method with DPC~\cite{InsafutdinovD18} and DRWR~\cite{handrwr2020} in terms of degeneration in Table~\ref{table:number2}. The degeneration of each method is the difference between the results obtained with $32^2$ and $128^2$ images. The least degeneration shows that the 2D projection matching can help our method to become more robust to the resolution of 2D supervision, Our degeneration is shown in Fig.~\ref{fig:Comparison2}.

\begin{table}[h]
\centering
\caption{Robustness to 2D supervision resolution.}  
\resizebox{0.7\linewidth}{!}{
    \begin{tabular}{c|c|c|c}  
     \hline
      Resolution & DPC & DRWR & Ours\\   
     \hline
       $32^2$ &5.01&3.95&3.41\\
       $128^2$ &3.62&3.49&3.10\\
     \hline
     Degeneration&1.39&0.46&\textbf{0.31}\\
     \hline
   \end{tabular}}
   \label{table:number2}
   \vspace{-0.28in}
\end{table}

\section{Conclusion}
We introduce a novel perspective to learn to generate fine structures for 3D point clouds in an unsupervised way. Current differentiable renderers depend on the per-pixel difference to infer 3D structures from 2D supervision, which however cannot fully capture 3D structures, especially for fine structures. Our method successfully resolves this issue by casting this problem into a 2D projection matching problem. By discretizing the continuous area covered by the silhouette into irregular point supervision, our method effectively pushes the neural network to learn to generate 3D point clouds whose 2D projections can match the irregular point supervision as accurately as possible. We also demonstrate that the irregular point supervision can reveal more specific structure information to learn, especially for fine 3D structures. Our outperforming experimental results show that our method can significantly improve the structure generation performance for 3D point clouds.

{\small
\bibliographystyle{ieee_fullname}
\bibliography{paper}
}

\end{document}